\documentclass[11pt]{article}

\usepackage[preprint]{acl}

\usepackage{times}
\usepackage{latexsym}

\usepackage[T1]{fontenc}
\usepackage{mdframed}
\usepackage{multirow}
\usepackage{colortbl}

\usepackage[utf8]{inputenc}

\usepackage{microtype}

\usepackage{subcaption}

\usepackage{inconsolata}

\usepackage{graphicx}

\usepackage{microtype}
\usepackage{hyperref}
\usepackage{url}
\usepackage{booktabs}
\usepackage{tabularx}

\usepackage{graphicx} 
\usepackage{amssymb}  
\usepackage{array}
\usepackage{xcolor}
\usepackage{soul}

\usepackage{longtable}
\usepackage{booktabs}
\usepackage{array}
\usepackage{ragged2e}

\usepackage{fvextra}

\usepackage{enumitem}
\usepackage{titlesec}

\definecolor{softhighlight}{rgb}{0.9, 0.95, 1.0} 
\sethlcolor{softhighlight}

\definecolor{darkblue}{rgb}{0, 0, 0.5}
\hypersetup{colorlinks=true, citecolor=darkblue, linkcolor=darkblue, urlcolor=darkblue}

\title{Generating Edit-Inducing Questions for AI Research Manuscripts}

\author{Sebastian Joseph$^{1}$\thanks{Work done during an internship at Adobe Research}\ \ \ \
Zichao Wang$^{2}$\ \ \ \
Jennifer Healey$^{2}$\ \ \ \ 
Alexa Siu$^{2}$\\
\textbf{Junyi Jessy Li}$^{1}$\ \ \ \ 
\textbf{Ani Nenkova}$^{2}$
\\
$^1$The University of Texas at Austin,
$^2$Adobe Research \\
{\small \tt \{sebaj,jessy\}@utexas.edu, \{jackwa,jehealey,asiu,nenkova\}@adobe.com} 
}

\begin{document}
\maketitle
\begin{abstract}
 
We study the ability of LLMs to generate \emph{edit-inducing questions} whose answer will improve a paper draft. 
On a dataset of paired submission and camera-ready papers from ICLR and NeurIPS, we compare the helpfulness of questions from GPT models with or without full paper context to that  of human reviewers.  
GPT produces more edit-inducing questions and its questions are associated with more extensive edits and cover a broader range of edited content compared to questions from reviewers. However, a much smaller percentage of the GPT questions are edit-inducing.
Our analyses confirm that automated questions can be beneficial to authors and highlight an example task where proper attending to long context deteriorates reasoning model ability to produce helpful output.

\end{abstract}

\section{Introduction}

Ineffective scientific writing is often due to the curse of knowledge: the authors know a lot about their problem and the details of their work but find it hard to imagine that their reader does not have this knowledge \cite{pinker2015sense}. A practical remedy is to request feedback on drafts, to find out what questions an intelligent and curious reader may have about the presented content. 

In academic practice, peer reviewers are often expected to provide constructive feedback to help authors improve the work itself as well as its presentation, in addition to their evaluation of novelty, soundness and significance. With the exponential growth of academic publications straining the peer review process, even this delayed feedback is harder to obtain~\cite{ stelmakh2020novicereviewerexperimentaddressscarcity, 10.1162/qss_a_00327, kim2025positionaiconferencepeer}. 
This reality has motivated a line of work on automatic feedback generation, enabled by peer review datasets \cite{kang2018datasetpeerreviewspeerread}: from early knowledge-graph and template-based systems \cite{Wang_2020} to LLM-based approaches spanning error detection \cite{liu2023reviewergptexploratorystudyusing}, large-scale empirical validation \cite{liang2023largelanguagemodelsprovide}, and multi-agent feedback generation \cite{darcy2024margmultiagentreviewgeneration}. Most recent work has shown that peer reviews can be mined to extract data for training models specifically designed to provide actionable feedback, with no evaluative elements. Automatic feedback has been deployed at scale and found useful and actionable by authors and organizers alike \cite{biswas2026aiassistedpeerreviewscale}.

\begin{figure}
    \centering
    \includegraphics[width=\linewidth]{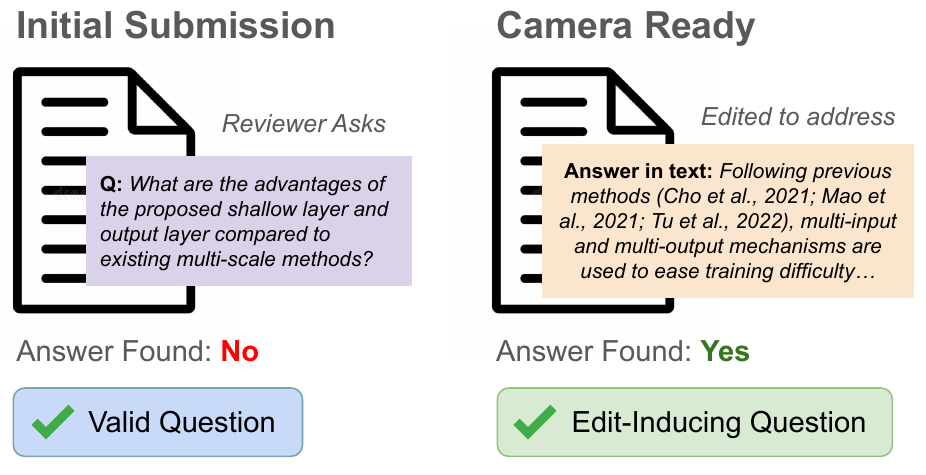}
    \caption{We seek to generate \textit{edit-inducing questions} like the example shown above. A question is asked based on content in the initial submission. When answered only in the camera ready version, we can expect content to have been edited to address such a question.}
    \label{fig:main_fig}
\end{figure}

Our interest is also in feedback generation, particularly \emph{in the form of questions} over a current draft.
Specifically we ask: \textit{(i) are the questions asked by human reviewers and LLMs edit-inducing? (ii) what fraction of author edits do such questions account for?} In other words, we quantify objectively through author decisions the quality of questions generated by LLMs and humans.

Our analyses are carried out on pairs of initial submission and final version of conference papers. By comparing the two versions, we pinpoint what content was added by the authors, presumably to improve the paper. Furthermore we measure whether questions asked by human peer reviewers, from the PeerQA dataset~\cite{baumgärtner2025peerqascientificquestionanswering}, and questions generated by LLMs target that content by searching for their answers in the papers. 

We find that LLMs have real potential as a source of edit-inducing questions. Most notably and unsurprisingly, they generate an order of magnitude more edit-inducing questions than the human reviewers.  
As a result of the increased volume of questions, their answers account for a larger percentage of edits made by the authors between their submitted and final paper versions.  We also find that model questions correspond to more extensive edits than questions from peer reviewers.

\section{Data \& Methodology}
We perform our analyses of edit-inducing questions on a subset of the PeerQA dataset \cite{baumgärtner2025peerqascientificquestionanswering}.

\paragraph{Data}
PeerQA contains questions extracted from peer reviews of machine learning and natural language processing papers, along with answers of these questions grounded in the final version of the paper. 
For 116 papers that were sourced from three venues hosted on OpenReview (ICLR 2022, ICLR 2023, NeurIPS 2022), we expanded the original dataset,  retrieving the initial submission of the paper and the final camera-ready version of the paper via the publicly accessible submission history. 
We extract the full text
using GROBID.

\paragraph{Question Generation}

Each paper is split into 512 token segments.
Figures and tables are considered their own paragraphs; segments are built by continually adding paragraphs until adding the next paragraph would exceed 512 tokens. A new segment is created at section boundaries to prevent text from multiple sections appearing in a single segment.
We pass each segment from the submitted version to GPT-4o and o3 with the prompt shown in the Appendix~\ref{sec:qgen_prompt}, with varying levels of context, to obtain insightful questions that can arise while reading that segment \cite{wu-etal-2023-qudeval,wu-etal-2024-questions}. The three context settings we evaluate here are: providing the full paper as context (\textbf{full}), providing every preceding segment to simulate linear reading (\textbf{linear}), and providing only the focused segment as context (\textbf{segment}).  

Table \ref{table:totalq} in Appendix~\ref{app:totalQ} shows the number of questions generated by each model+context combination, along with the number of questions from peer reviews of the same papers. On average, there are three questions from peer reviewers per paper; the automatically generated questions are many more, ranging from about 140 on average per paper for GPT-4o segment to 206 for o3 segment.

\paragraph{Answer Detection}

For each question, we use the approach from the PeerQA paper to find if an answer exists in the submitted and camera-ready versions of the paper. 
We use the OpenAI \texttt{text-embedding-small} to find the five paper segments most similar to the question. 
Then we prompt \texttt{gpt-4o} (see Appendix~\ref{app:answer_prompt}) to decide if any of these five segments answers the question.

\section{Measuring Edit-Inducing Capability}

We define four measures of question helpfulness, with higher values indicating more helpful properties. These measures were calculated per paper and are presented as an average over all papers.

\paragraph{Valid questions} Both peer reviewers and GPT ask questions that are already answered in the submitted manuscript. We call a question \emph{valid} if it is not already answered in the submitted manuscript. The percentage of valid questions reflects the ability of the LLM to take the context of the entire paper into account when asking a question.

\paragraph{Edit-inducing questions} Edit inducing questions are a subset of valid questions. These are valid questions that the authors answer in the final version. Assuming all changes are made with the intention to improve the paper, edit-inducing questions are empirically verified to have value as actionable feedback on a manuscript draft. The valid questions that are not edit-inducing remain unanswered. Appendix~\ref{app:q_examples} presents randomly drawn edit-inducing and unanswered questions. An auxiliary measure is "Edit-inducing answer rate", equal to the percentage of answered questions that are answered only in the final version.

\paragraph{Edit rate} characterizes a single edit-inducing question. It measures the extent to which the text in the 
related segment is changed to provide the answer. For an edit-inducing question
and a corresponding segment from the final submission that contains its answer, it equals to the proportion of \textit{edited} (added or substituted) characters in that segment with respect to the total number of characters that were originally in the segment of the submitted manuscript.
Appendix \ref{app:editRateExamples} shows examples of segment changes and the respective edit rate values.

\paragraph{Edit coverage} characterizes a set of edit-inducing questions, here the set of all edit-inducing questions generated by a given source. It equals the fraction of edited characters  attributable to all edit-inducing
questions out of all characters that changed between the two drafts. Its value reflects the extent to which the answers to all edit-inducing questions cover the entire content authors found worth adding to their papers.

\section{Results}

\begin{figure*}
    \centering
    \includegraphics[width=1\linewidth]{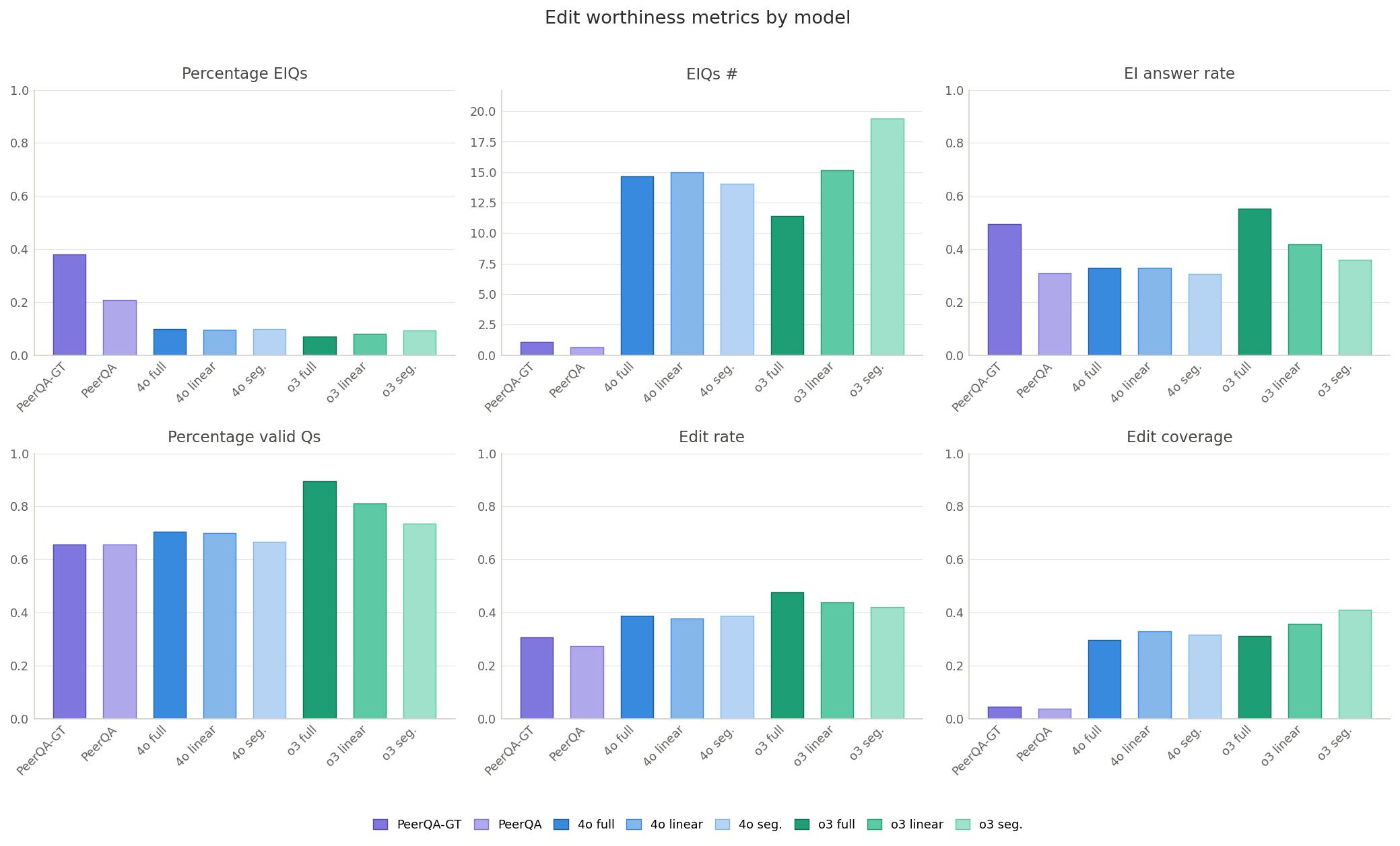}
    \caption{Measures of edit worthiness of questions asked by peer reviewers (PeerQA) and GPT models.}
    \label{fig:bar_charts}
\end{figure*}

Figure~\ref{fig:bar_charts} shows the quality metrics for the sets of questions generated from peer reviews, and by GPT-4o and o3 with different levels of context.

\subsection{Human Peer-Review Questions}

We present two evaluations of the human questions from the PeerQA dataset. They vary in how answer detection is done for the camera-ready version. In the PeerQA-GT setting, the answer is found in the paper based on the ground truth, human annotated answers in the dataset. 

The other setting,  PeerQA, we apply  the automated answer detection pipeline. 
Both settings used the automated pipeline to find answers in the initial submission as there is no corresponding ground truth.
Both show similar results, albeit with the automated pipeline having weaker recall of answered questions.

Around two-thirds (65\%) of the peer review questions are not answered in a submitted paper, in other words 35\% of the human questions were already answered in the submitted version. 

Roughly one-third (38\%) of human peer-review questions for a paper are edit inducing. In number of questions, this translates on average to around 1 such question per paper. Therein lies the primary issue with these questions, which is their scarcity. This small amount of questions also explains the low edit coverage, with edit-inducing questions from peer reviews covering, on average, only around 4\% of all edits made in the paper. Authors, in their edits, incorporate a lot more feedback beyond what was addressed by their peer reviews.

\subsection{GPT Edit-Inducing Questions}
Type of model and context make minor differences in metrics. Overall either setting can be used and all compare similarly to human questions. The GPT models generate markedly more total questions, and, as a result, generate a magnitude more edit-inducing questions, between 11 and 19 questions per paper. However, as a percentage, only around 7-10\% of all questions for a paper were edit-inducing. The models over-generate questions, lacking strong discriminative capability to restrict themselves from asking questions that lack utility.

The models also ask more valid questions as a percentage compared to human peer-reviewers. They seem to have a better ability to recall from the context the type of questions that could already be answered by the paper. 
The non-reasoning model has a minor edge over the human question, and barely any difference in percentage valid questions with varying context. The reasoning model however makes good use of the paper context and generates considerably larger percentage of valid questions than humans do, when provided with the full paper as context. 
Larger context settings also tend to yield a higher proportion of valid questions. However, even when given only a document segment, models are capable of asking questions that are unanswerable by the rest of the paper. 

Edit inducing answer rate numbers closely track the percentage valid questions. It is a redundant measurement, with all conclusions aligned with those from percentage valid. 

The edit rate for GPT questions is higher than that for human questions, indicating the questions correspond to larger spans of edited content. 
The edit coverage of GPT questions is much higher than that for peer questions. GPT generates larger number of edit inducing questions capable of targeting a wider range of improvements.

\subsection{Copying Behavior}
\label{sec:copying}

In terms of edit-inducing ability, the questions generated from the reasoning model appear very much on par with those from the non-reasoning model. However in qualitative inspection, as in the samples shown in Appendix~\ref{app:q_examples}, we noticed that 4o questions appear more extractive, often paraphrasing the content being discussed in the form of a question and repeating many words from the segment for which the questions were generated. In that respect, they appear to be worse than those generated by o3. To quantify the difference, we compute the average proportion of words in a question that overlap with words in the span of text from the paper that elicited that question.

\begin{table}[ht]
\centering
\begin{tabular}{lc}
\hline
\textbf{Name} & \textbf{Overlap Proportion} \\
\hline
o3                 & 0.2337 \\
o3 Linear          & 0.2337 \\
o3 Segment     & 0.2306 \\
GPT-4o             & 0.3020 \\
GPT-4o Linear      & 0.3069 \\
GPT-4o Segment & 0.3337 \\
\hline
\end{tabular}
\caption{Proportion of words in the question overlapping with words found from the span of text that elicited it}
\label{tab:q_span_rat}
\end{table}

Table~\ref{tab:q_span_rat} shows the results for each model and context combination. 
We see a significant increase in this word overlap with questions generated by GPT-4o, indicating a higher degree of extractive behavior in how the model develops its questions. 

\section{Discussion and Future Work}

A key issue that we highlighted with LLM-generated questions was the low proportion of edit-inducing questions. Authors would likely prefer to engage with a higher rate of such questions. Pruning down the generated questions effectively and cheaply remains an open problem for future work.
We made the assumption that questions that are edit-inducing are more preferable for feedback. A future experiment with human subjects may probe the reasons why certain questions were not answered. It is possible that there were good questions that were not answered through edited content because manuscript page limits prevent authors from including all details, or because they did not have time to answer an otherwise good question.
This paper highlights the concept of edit-inducing questions as a signal for useful feedback. Ultimately, developing a single model capable of generating all edit-inducing questions would be ideal for practical feedback to authors.

\section{Conclusion}

We present a study that confirms that large language models may benefit writers of papers in the artificial intelligence area, by serving as an indefatigable reader, asking questions that arise from reading segments of the paper. Models produce a order of magnitude more questions than peer reviewers, with a much larger number of edit-inducing questions. Individual questions from LLMs have higher edit-rates than peer questions, and all automatically produced questions account for about half of all the edits that authors made between their initial submission and the final version. Reasoning models are better at generating valid questions, even when not taking the full paper as context when producing the questions; their output feels qualitatively better, objectively less extractive in wording than non-reasoning models. 

\section*{Limitations}

Our experiments are based on a single family of closed models, GPT. However, we were able to vary context and type of model for a complete set of experiments. Inclusion of more models would have resulted in overwhelming tables. Choosing one model makes it easy to see that most of the modeling choices make no difference. 

The assessment of edit inducing capability in this paper utilizes automated answer detection extensively. 
However, as we show in Appendix~\ref{sec:ans_f1}, this automated system is effective in detecting answers, especially in terms of precision. Thus results we present lead to reasonable estimates of the potential utility from LLM generated questions. The comparison with gold standard human labels of answers indicates that if anything, the percentage of edit inducing questions generated from LLMs may be higher in practice than the ones we report here. 

Because of the constraints of using the PeerQA dataset, we only looked at papers that were accepted in the same venue that they were submitted in. Applying a similar analysis over rejected papers that were accepted in a future venue would be of interest for future work.

\bibliography{custom}

@misc{liang2023largelanguagemodelsprovide,
  title = {Can large language models provide useful feedback on research papers? A large-scale empirical analysis},
  author = {Weixin Liang and Yuhui Zhang and Hancheng Cao and Binglu Wang and Daisy Ding and Xinyu Yang and Kailas Vodrahalli and Siyu He and Daniel Smith and Yian Yin and others},
  year = {2023},
  eprint = {2310.01783},
  archiveprefix = {arXiv},
  primaryclass = {cs.LG},
  url = {https://arxiv.org/abs/2310.01783},
  journal = {NEJM AI},
  doi = {10.48550/arXiv.2310.01783},
}

@book{pinker2015sense,
  title = {The sense of style: The thinking person's guide to writing in the 21st century},
  author = {Pinker, Steven},
  year = {2014},
  publisher = {Penguin Books},
  doi = {10.5860/choice.191126},
}

@inproceedings{wu-etal-2024-questions,
  title = {Which questions should {I} answer? Salience Prediction of Inquisitive Questions},
  author = {Wu, Yating  and Mangla, Ritika  and Dimakis, Alexandros G.  and Durrett, Greg  and Li, Junyi Jessy},
  editor = {Al-Onaizan, Yaser  and Bansal, Mohit  and Chen, Yun-Nung},
  booktitle = {Conference on Empirical Methods in Natural Language Processing},
  year = {2024},
  journal = {Conference on Empirical Methods in Natural Language Processing},
  pages = {19969-19987},
  doi = {10.48550/arXiv.2404.10917},
  publisher = {Association for Computational Linguistics},
}

@inproceedings{wu-etal-2023-qudeval,
  title = {{QUD}eval: The Evaluation of Questions Under Discussion Discourse Parsing},
  author = {Wu, Yating  and Mangla, Ritika  and Durrett, Greg  and Li, Junyi Jessy},
  editor = {Bouamor, Houda  and Pino, Juan  and Bali, Kalika},
  booktitle = {Conference on Empirical Methods in Natural Language Processing},
  year = {2023},
  journal = {Conference on Empirical Methods in Natural Language Processing},
  pages = {5344-5363},
  doi = {10.48550/arXiv.2310.14520},
  publisher = {Association for Computational Linguistics},
}

@misc{baumgärtner2025peerqascientificquestionanswering,
  title = {PeerQA: A Scientific Question Answering Dataset from Peer Reviews},
  author = {Tim Baumgärtner and Ted Briscoe and Iryna Gurevych},
  year = {2025},
  eprint = {2502.13668},
  archiveprefix = {arXiv},
  primaryclass = {cs.CL},
  url = {https://arxiv.org/abs/2502.13668},
  journal = {North American Chapter of the Association for Computational Linguistics},
  pages = {508-544},
  doi = {10.48550/arXiv.2502.13668},
  publisher = {Association for Computational Linguistics},
}

@misc{darcy2024margmultiagentreviewgeneration,
  title = {MARG: Multi-Agent Review Generation for Scientific Papers},
  author = {Mike D'Arcy and Tom Hope and Larry Birnbaum and Doug Downey},
  year = {2024},
  eprint = {2401.04259},
  archiveprefix = {arXiv},
  primaryclass = {cs.CL},
  url = {https://arxiv.org/abs/2401.04259},
  journal = {arXiv.org},
  doi = {10.48550/arXiv.2401.04259},
}

@misc{biswas2026aiassistedpeerreviewscale,
  title = {AI-Assisted Peer Review at Scale: The AAAI-26 AI Review Pilot},
  author = {Joydeep Biswas and Sheila Schoepp and Gautham Vasan and Anthony Opipari and Arthur Zhang and Zichao Hu and Sebastian Joseph and Matthew Lease and Junyi Jessy Li and Peter Stone and others},
  year = {2026},
  eprint = {2604.13940},
  archiveprefix = {arXiv},
  primaryclass = {cs.AI},
  url = {https://arxiv.org/abs/2604.13940},
}

@misc{kang2018datasetpeerreviewspeerread,
  title = {A Dataset of Peer Reviews (PeerRead): Collection, Insights and NLP Applications},
  author = {Dongyeop Kang and Waleed Ammar and Bhavana Dalvi and Madeleine van Zuylen and Sebastian Kohlmeier and Eduard Hovy and Roy Schwartz},
  year = {2018},
  eprint = {1804.09635},
  archiveprefix = {arXiv},
  primaryclass = {cs.CL},
  url = {https://arxiv.org/abs/1804.09635},
  journal = {North American Chapter of the Association for Computational Linguistics},
  pages = {1647-1661},
  doi = {10.18653/v1/N18-1149},
  publisher = {Association for Computational Linguistics},
}

@inproceedings{Wang_2020,
  title = {ReviewRobot: Explainable Paper Review Generation based on Knowledge Synthesis},
  url = {http://dx.doi.org/10.18653/v1/2020.inlg-1.44},
  doi = {10.18653/v1/2020.inlg-1.44},
  booktitle = {International Conference on Natural Language Generation},
  publisher = {Association for Computational Linguistics},
  author = {Wang, Qingyun and Zeng, Qi and Huang, Lifu and Knight, Kevin and Ji, Heng and Rajani, Nazneen Fatema},
  year = {2020},
  pages = {384–397},
  journal = {International Conference on Natural Language Generation},
}

@misc{liu2023reviewergptexploratorystudyusing,
  title = {ReviewerGPT? An Exploratory Study on Using Large Language Models for Paper Reviewing},
  author = {Ryan Liu and Nihar B. Shah},
  year = {2023},
  eprint = {2306.00622},
  archiveprefix = {arXiv},
  primaryclass = {cs.CL},
  url = {https://arxiv.org/abs/2306.00622},
  journal = {arXiv.org},
  doi = {10.48550/arXiv.2306.00622},
}

@article{10.1162/qss_a_00327,
  author = {Hanson, Mark A. and Barreiro, Pablo Gómez and Crosetto, Paolo and Brockington, Dan},
  title = {The strain on scientific publishing},
  journal = {Quantitative Science Studies},
  volume = {5},
  number = {4},
  pages = {823-843},
  year = {2023},
  month = {11},
  issn = {2641-3337},
  doi = {10.1162/qss_a_00327},
  url = {https://doi.org/10.1162/qss_a_00327},
  eprint = {https://direct.mit.edu/qss/article-pdf/5/4/823/2478590/qss_a_00327.pdf},
}

@misc{stelmakh2020novicereviewerexperimentaddressscarcity,
  title = {A Novice-Reviewer Experiment to Address Scarcity of Qualified Reviewers in Large Conferences},
  author = {Stelmakh, Ivan and Shah, Nihar B. and Singh, Aarti and Daum'e, Hal},
  year = {2020},
  eprint = {2011.15050},
  archiveprefix = {arXiv},
  primaryclass = {cs.HC},
  url = {https://arxiv.org/abs/2011.15050},
  journal = {AAAI Conference on Artificial Intelligence},
  doi = {10.1609/aaai.v35i6.16610},
}

@misc{kim2025positionaiconferencepeer,
  title = {Position: The AI Conference Peer Review Crisis Demands Author Feedback and Reviewer Rewards},
  author = {Jaeho Kim and Yunseok Lee and Seulki Lee},
  year = {2025},
  eprint = {2505.04966},
  archiveprefix = {arXiv},
  primaryclass = {cs.AI},
  url = {https://arxiv.org/abs/2505.04966},
  journal = {International Conference on Machine Learning},
  doi = {10.48550/arXiv.2505.04966},
}

\appendix

\onecolumn

\section{Question Generation Prompt}
\label{sec:qgen_prompt}

\begin{Verbatim}[
    frame=single, 
    framesep=3mm, 
    label=Prompt A, 
    fontfamily=courier, 
    fontsize=\tiny,
    breaklines=true,
    breakanywhere=true,
    breaksymbol={},
    breaksymbolleft=,  
    breaksymbolright=  
]

System: You are logical, intelligent, insightful, precise, and can understand the contents of research papers. You are knowledgeable on different fields and domains of science and engineering. You are able to interpret research papers, create questions and answers, and compare multiple aspects.

Imagine the following scenario: You are reviewing a submitted research paper manuscript. After you are done reading, you have questions you want to ask the authors. You are a very intelligent reader so you don't ask questions that are already answered within the paper.

I will provide you two things: First, context which you have already read, and a chunk of text that you are currently reading.

This is your task. First, you need to look at the context that is provided to you. This is information that you have already read and aware of. Please don't ask questions that can be answered through this context. What you need to do instead is generate all possible follow-up questions arising from the provided chunk of text that you are currently reading. These follow-up questions must be insightful and highlight a critical gap of information that a reader would desire to know while reading.

You MUST follow these rules when creating these questions:
(1) The question must not be answered by text in the context. Make sure to thoroughly read the text in the context and craft a question that cannot be answered by it.
(2) These questions must be questions that a smart reader is naturally thinking about when reading this chunk of text. Remember, they also have already read the context as well, so they will not be thinking of questions that they already know the answer to. Put yourself in this smart reader's shoes when crafting these questions.

I will provide you with a typology of these questions below. You can use this typology to come up with more diverse questions.

Typology:
  - Content: 
    - content-clarification: questions that try to clarify something that you've read
  - Insights:
      - insights-more: questions that ask for more insights
      - insights-nuance: questions that ask for more nuance in the insights
      - insights-soundness: questions that ask whether the insight is sound.
  - Measurement:
      - measurement-more: questions that ask for more metrics to evaluate in experiments
      - measurement-detail: questions that ask for more detail about some measurement/metrics
      - measurement-alternative: questions that ask why an alternative measurement wasn't used
  - Method: 
      - method-alternative: questions that probe at an alternative method that has a similar effect
      - method-detail: questions that probe for more detail about the method
      - method-motivation: questions that probe at the motivation for using such a method

When you respond, first think really hard, step-by-step, about every span of text in the provided chunk, and think about what would be good, thoughtful, and interesting follow-up questions according to the above rules.

Then you need to provide me with these follow-up questions and also tell me the exact span of text each follow-up question corresponds to, as in follows up on, in your response. Absolutely make sure that the span you provide is an exact substring of the chunk text, and that it is not a summary or paraphrase. It should be the exact text corresponding to the question.

If you can't come up with any follow-up questions at all (which is perfectly fine), just return None.

### REASONING PROCESS
  <Explain your reasoning and thinking process>

### JSON
  { 
      "follow_up_questions": [
                          {
                              "question": "<generated follow-up question>",
                              "span": "<exact span of text in the provided chunk corresponding to this question>"
                              "question_type": <Content/Insights/Measurement/Method>,
                              "question_subtype": "<content-clarification/insights-more/insights-nuance/insights-soundness/measurement-more/measurement-detail/measurement-alternative/method-alternative/method-detail/method-motivation>"
                          }, 
                          
                          ...
                      ] or None
  }

User: 

[Start of Context]
{context}
[End of Context]

[[Start of *Provided Chunk*]]
{segment}
[[End of *Provided Chunk*]]

\end{Verbatim}

\section{Answerability Prompt}
\label{app:answer_prompt}

\begin{Verbatim}[
    frame=single, 
    framesep=3mm, 
    label=Prompt B, 
    fontfamily=courier, 
    fontsize=\tiny,
    breaklines=true,
    breakanywhere=true,
    breaksymbol={},
    breaksymbolleft=,  
    breaksymbolright=  
]

System: You are logical, intelligent, precise, and can understand the contents of research papers. You are knowledgeable on different fields and domains of science and engineering. You are able to interpret research papers, and determine whether information inside these papers can answer some question.

This is your task: Read the following several chunks from a paper and answer whether the question can be answered by one or more of these chunks. If the question can be answered, report the chunk numbers of the chunks where you found the answer and answer the question in a sentence or two using only the information in that chunk(s).

When you respond, first think really hard, step-by-step, about every sentence in every chunk, and analyze whether this information can answer the question provided. 

You can definitely find the question to be unanswerable. You can only find a question to be answerable if ONLY the information within the provided chunks fully answers this question.

If you find the question to be answerable, please select the fewest possible chunks you would need to answer the question. If there is one chunk that can fully answer this question, please select only that chunk. If there are multiple chunks that can fully answer this question, please select only the one chunk you think best answers the question. You should only select multiple chunks if the only way to fully answer the question is by combining the information in these chunks.

The chunk numbers are 1-indexed, meaning the first chunk is chunk number 1, the second chunk is chunk number 2, and so on. When you recieve the chunks, you will know the chunk number as it appears like this: "CHUNK #<chunk_number>". Please only report the exact number, and only the number, of the chunks you selected. Do not report a chunk number that is not in the range of possible chunk numbers. The chunk number **should not be less than 1** nor should it be greater than the maximum chunk number. This is unforgivable. 

In addition to selecting the chunks, you also need to provide the exact span of text within the chunk that answers the question. This span should be a substring of the chunk text that directly addresses the question. Absolutely make sure that the span you provide is an exact substring of the chunk text, and that it is not a summary or paraphrase. It should be the exact text that answers the question.

You need to provide me with an binary Yes or No as the answer to whether the question is answerable by the provided chunks. If answerable, provide me with the list of chunk numbers corresponding to your selected chunks. I also require you detail the exact span of text within the chunk that answers the question. If unanswerable, this can be an empty list.

If answerable, provide me with a 1-2 sentence answer. Otherwise, just answer with an empty string.


### REASONING PROCESS
<Explain your reasoning and thinking process>

### JSON
{
  "is_answerable": "<Yes/No>",
  "selected_chunks": [
    {
      "chunk_number": <chunk number>,
      "span": "<exact span of text within the chunk that answers the question>"
    },
    ...
  ],
  "answer": "<1-2 sentence answer, or empty string if unanswerable>"
}

User: 

[Start of Provided Chunks]
{answer_batch}
[End of Provided Chunks]

[Start of Question]
{question}
[End of Question]

\end{Verbatim}

\section{Total number of generated questions}
\label{app:totalQ}

Table \ref{table:totalq} Lists the total, average, minimum and maximum number of questions generated by each source. Regardless of model and context, GPT generate many more questions than peer reviewers do. 4o and o3 exhibit a different pattern as the context length changes. 4o produces fewer questions when presented only with the segment, compared to having the full paper as context as well. On average, the full paper context yields 15 more questions per paper. The reverse trend happens for o3: it produces 206 question on average when presented with the segment only but only 163 questions when the full paper is provided as well. The variation of number of questions is much bigger for o3, 40 question difference on average. 

It is worth noting that the papers themselves modulate the number of produced questions. There is a tenfold difference in the number of questions for the paper with fewest questions compared to that with most questions. 

\begin{table}
    \centering
\begin{tabular}{lcccc}
\toprule
\textbf{Model} & \textbf{\shortstack{Total\\Qs}} & \textbf{\shortstack{Avg Qs/\\Paper}} & \textbf{\shortstack{Max\\Qs}} & \textbf{\shortstack{Min\\Qs}} \\
\midrule
PeerQA      & 362    & 3.12   & 11  & 1  \\ \midrule
4o full     & 17,734 & 152.88 & 320 & 38 \\
4o linear    & 18,158 & 156.53 & 321 & 32 \\
4o segment   & 16,018 & 138.09 & 416 & 28 \\ \midrule
o3 full     & 18,940 & 163.28 & 433 & 47 \\
o3 linear    & 21,339 & 183.96 & 455 & 43 \\
o3 segment   & 23,906 & 206.09 & 586 & 45 \\
\bottomrule
\end{tabular}
\caption{Number of questions by source.}
\label{table:totalq}
\end{table}

\section{Assessing Answerability}
\label{sec:ans_f1}

Table \ref{tab:f1_results} shows the precision and recall of the automated answerability pipeline evaluated on the PeerQA gold standard data. Precision is quite good but an existing answer is often not detected. 

As for five segment cutoff, providing the entire document to the answerability model to answer every question we evaluate is prohibitively expensive. Providing only the top 5 segments proved to be a worthy tradeoff to ensure experiments stayed within budget while capturing the vast majority of existing answers. 
Preliminary tests showed that the ground truth answer was contained in the top 5 answer segments 72\% of the time, which was reasonable enough to us to believe in the reliability of this pipeline under such a cutoff.

\begin{table}[h]
    \centering
    \small
    \begin{tabular}{lc}
        \toprule
        Metric & Score \\
        \midrule
        Precision & 0.836 \\
        Recall & 0.586 \\
        F1 & 0.689 \\
        \bottomrule
    \end{tabular}
    \caption{Precision, recall, and F1 @ 5 (among the top 5 answers) for \texttt{gpt-4o} deciding if a segment is an answer to a given question. Evaluated against ground truth PeerQA answers.}
    \label{tab:f1_results}
\end{table}

\section{Edit-inducing and unanswered questions}
\label{app:q_examples}

Table~\ref{tab:examples_table} below shows random examples of edit-inducing and unanswered questions produced by different model and context settings.

\newlength{\colw}
\setlength{\colw}{\dimexpr(\linewidth - 2\tabcolsep - \arrayrulewidth)/2\relax}
 
\begin{longtable}{@{}p{\colw}|p{\colw}@{}}
\caption{Side-by-side comparison of edit-inducing and unanswered questions.}
\label{tab:examples_table}
\\
\toprule
\textbf{Edit-Inducing} & \textbf{Unanswered} \\
\midrule
\endfirsthead
\multicolumn{2}{c}{\tablename~\thetable{} \emph{(continued)}} \\[4pt]
\toprule
\textbf{Edit-Inducing} & \textbf{Unanswered} \\
\midrule
\endhead
\bottomrule
\endlastfoot
 
\multicolumn{2}{l}{\small\textit{Model: \textbf{o3}}} \\
\midrule
\textsc{\footnotesize Method} {\footnotesize\textit{detail}} \newline \small Could you explicitly show how a single layer of a k-GNN is encoded as a TL\textasciicircum{}(t)\_\{k+1\}(?) expression, in particular how the higher--order neighborhood aggregation and the permutation--invariant updates are handled within only k+1 index variables? & \textsc{\footnotesize Method} {\footnotesize\textit{detail}} \newline \small FIND\_HP requires its input point to be $\delta$-non-degenerate, yet Step 1 only guarantees that x\_i is a critical point.  What mechanism or additional test ensures that each x\_i returned by FIND\_CP actually satisfies the $\delta$-non-degeneracy condition needed by FIND\_HP? \\
\textsc{\footnotesize Insights} {\footnotesize\textit{nuance}} \newline \small Why do poisoned and benign frames stay close in the SiamFC++ feature space whereas they separate in SiamFC and SiamRPN++?  Could you elaborate on what architectural or training differences lead to this behavior? & \textsc{\footnotesize Method} {\footnotesize\textit{detail}} \newline \small You refer to an additional invariant layer that has a "similar (simpler) representation as given in equation 1 (Maron et al., 2019c)".  Could you provide the explicit TL representation of this invariant layer and explain why it is simpler? \\
\textsc{\footnotesize Method} {\footnotesize\textit{detail}} \newline \small For the min-max and z-score baselines, are the normalization statistics computed per instance, per batch, or over the whole training set, and are they recomputed during inference? & \textsc{\footnotesize Method} {\footnotesize\textit{detail}} \newline \small How exactly are the adversarially generated `confusing samples' produced (e.g., number of optimization steps, step size, loss function, and whether this generation is performed on-the-fly during training or pre-computed)? \\
\textsc{\footnotesize Insights} {\footnotesize\textit{soundness}} \newline \small On what empirical or theoretical grounds do you characterise $\delta$-regularity as a ``mild'' general position assumption?  Have you measured how frequently trained networks satisfy $\delta$-regularity without the artificial perturbation you propose? & \textsc{\footnotesize Method} {\footnotesize\textit{detail}} \newline \small The sentence states that edge weights depend on both distance and direction, yet the Gaussian kernel you provide only uses the squared distance. How is the direction information actually incorporated into the adjacency weights? \\
\textsc{\footnotesize Measurement} {\footnotesize\textit{more}} \newline \small Can you provide ablation results quantifying the benefit of the parallel exploration relative to a single-process baseline to substantiate the claim of "efficient exploration"? & \textsc{\footnotesize Method} {\footnotesize\textit{detail}} \newline \small You claim HyperDQN measures uncertainty through "infinite" ensembles, but in implementation only finitely many z-samples can be used. How many z vectors are actually drawn per training step and per episode, and how did you decide on this number? \\
\textsc{\footnotesize Insights} {\footnotesize\textit{nuance}} \newline \small Remark 1 suggests $\beta$\textsubscript{k}=0 causes stagnation; is there theoretical guidance on how small $\beta$\textsubscript{k} can safely be before convergence degrades, and how does this threshold relate to the contractivity constant of g? & \textsc{\footnotesize Measurement} {\footnotesize\textit{more}} \newline \small Could you also report confidence intervals for the performance differences to complement the p-values and provide an estimate of effect size? \\
\textsc{\footnotesize Method} {\footnotesize\textit{detail}} \newline \small Equation (18) requires the ratio d\_\{$\pi$*\_r\}(s,a)/d\_D(s,a), which itself depends on V* and ?r; could you clarify the computational procedure for obtaining this ratio at policy-extraction time, especially in high-dimensional continuous spaces? & \textsc{\footnotesize Method} {\footnotesize\textit{detail}} \newline \small Since Barlow Twins forces each dimension to have norm $\sqrt{}$N, does varying the batch size during training (e.g., in distributed settings) introduce instability or require re-scaling of other loss terms? \\
\textsc{\footnotesize Measurement} {\footnotesize\textit{detail}} \newline \small For the Sensitivity-n experiments in Figures 10 and 11, what exact subset sizes (n or percentage of pixels) were evaluated, and how many random subsets were sampled for each n? & \textsc{\footnotesize Method} {\footnotesize\textit{detail}} \newline \small How exactly is the `identity' initialization implemented for the depth-wise convolution inserted between grown convolutional layers (e.g., kernel size, padding, and channel mapping)? \\
\textsc{\footnotesize Method} {\footnotesize\textit{detail}} \newline \small What specific mechanisms (e.g., different initializations, data bootstrapping, parameter regularisation) do you employ to encourage sufficient diversity among the models in your ensemble whose variance is used for epistemic uncertainty estimation? & \textsc{\footnotesize Method} {\footnotesize\textit{detail}} \newline \small Since the output dimension of a depth-wise tensor product is tied to the input irreps dimensions, does this restrict your ability to expand channel capacity within a layer? Do you follow the DTP with any additional projections to compensate, and what is the computational impact? \\
\textsc{\footnotesize Content} {\footnotesize\textit{clarification}} \newline \small Lines 5--7 reuse the variable names x\_g and x'\_g for $\sigma$ = 2 and $\sigma$ = 3 Gaussian filters, apparently overwriting the first blurred images. Did you intend to keep **two distinct blurred versions** ($\sigma$ = 2 and $\sigma$ = 3), and if so could you clarify the correct variable naming and the computation order? & \textsc{\footnotesize Method} {\footnotesize\textit{detail}} \newline \small Will you publish a standardized data-collection protocol (camera height, angle, lighting conditions, annotation format, etc.) for external contributors to ensure new submissions remain compatible with the existing dataset? \\
\midrule
\multicolumn{2}{l}{\small\textit{Model: \textbf{o3 segment}}} \\
\midrule
\textsc{\footnotesize Measurement} {\footnotesize\textit{detail}} \newline \small Which concrete values of S (number of searches) and R (number of retrievals) were explored, and how does model performance vary across these individual settings? & \textsc{\footnotesize Method} {\footnotesize\textit{detail}} \newline \small What is the architecture and training configuration of the `challenging CIFAR10 network' on which you evaluated your method? \\
\textsc{\footnotesize Method} {\footnotesize\textit{detail}} \newline \small What regularization or early-stopping strategies do you employ during the meta-training phase to mitigate overfitting to the training tasks? & \textsc{\footnotesize Insights} {\footnotesize\textit{soundness}} \newline \small Have you investigated values of $\alpha$ other than 0.5 in the $\alpha$-energy objective, and how sensitive is model performance to this hyper-parameter? \\
\textsc{\footnotesize Measurement} {\footnotesize\textit{detail}} \newline \small The tighter bounds $\mu$ $\leq$ \~{O}($\varepsilon$\textasciicircum{}\{5/8\} d\textasciicircum{}\{1/4\}) and $\mu$ $\leq$ \~{O}($\varepsilon$\textasciicircum{}\{1/2\} d\textasciicircum{}\{1/4\}) are prescribed for Lines 6 and 8 of Algorithm 1; could you explain why different exponents on $\varepsilon$ are needed for these two lines and how sensitive the convergence guarantee is to these choices? & \textsc{\footnotesize Measurement} {\footnotesize\textit{more}} \newline \small Have you measured how frequently the zig-zag pattern actually appears versus disappearing for extremely hard samples across your datasets, and if so what proportion of samples follow each behaviour? \\
\textsc{\footnotesize Method} {\footnotesize\textit{detail}} \newline \small When applying the model trained on LVIS to COCO and Objects365, how do you reconcile duplicate or synonymous category names across datasets? & \textsc{\footnotesize Measurement} {\footnotesize\textit{detail}} \newline \small What exactly does the metric 'Consistency and Repetition' measure in the GSM8K dataset, and how should a value of 1 be interpreted by the reader? \\
\textsc{\footnotesize Method} {\footnotesize\textit{detail}} \newline \small How is the marginal utility u\_i(s, a\_i, a\_\{-i\}) actually estimated within your framework when the teammates' future actions are unknown or only partially observable? & \textsc{\footnotesize Content} {\footnotesize\textit{clarification}} \newline \small The enumeration ends with "3." but no model description follows---what is the third model in your zero-shot evaluation suite? \\
\textsc{\footnotesize Method} {\footnotesize\textit{detail}} \newline \small Since the rule body is said to correspond to a walk from E\textsubscript{1} to E\_\{l+1\}, do you impose any constraints (e.g., acyclicity, maximum length, unique entities) on this walk when mining or learning rules? & \textsc{\footnotesize Method} {\footnotesize\textit{detail}} \newline \small What concrete scale factors or resolutions are used in the multi-scale augmentation when collecting pseudo boxes, and how were these scales selected? \\
\textsc{\footnotesize Method} {\footnotesize\textit{detail}} \newline \small Could you formally describe the new initialization method you introduce, including how you set both the incoming and outgoing weights of the added neurons? & \textsc{\footnotesize Measurement} {\footnotesize\textit{detail}} \newline \small Which specific DARTS-based baseline methods are included in Table-1, and what evaluation metric (e.g., top-1 accuracy, error rate) is used for the comparison? \\
\textsc{\footnotesize Method} {\footnotesize\textit{alternative}} \newline \small In the special case where all three eigenvalues are equal (a perfectly spherical distribution), how is the coordinate frame selected and does the method still guarantee rotational equivariance? & \textsc{\footnotesize Measurement} {\footnotesize\textit{detail}} \newline \small Are the numbers of classes and images balanced across the Easy, Medium, and Hard bins, and what criteria determined the class counts per bin? \\
\textsc{\footnotesize Content} {\footnotesize\textit{clarification}} \newline \small What distribution does $\rho$ represent in the expectation E\_\{s0:$\infty$$\sim$$\rho$0:$\infty$\} and what is the meaning of the superscript G in $\rho$0:$\infty$\textasciicircum{}G? & \textsc{\footnotesize Method} {\footnotesize\textit{alternative}} \newline \small Why did you decide not to incorporate representation distillation (as in ViLD) or prompt optimization (as in DetPro) in F-VLM, and what trade-offs did you observe when experimenting with these alternatives? \\
\textsc{\footnotesize Content} {\footnotesize\textit{clarification}} \newline \small Could you clarify what M\_i represents in the repetition formula and how it is determined for variable-length reasoning steps? & \textsc{\footnotesize Method} {\footnotesize\textit{alternative}} \newline \small Have you considered using alternative approaches like temporal convolutional networks (TCNs) or transformer-based architectures instead of storing representative patterns, and if so, what were the comparative results? \\
\midrule
\multicolumn{2}{l}{\small\textit{Model: \textbf{GPT-4o segment}}} \\
\midrule
\textsc{\footnotesize Measurement} {\footnotesize\textit{detail}} \newline \small What criteria were used to select the smoothing parameter $\mu$, and how sensitive is the method to variations in this parameter? & \textsc{\footnotesize Method} {\footnotesize\textit{motivation}} \newline \small What motivates the assumption that the noisy gradient is bounded by M\_g in equation (84), and how robust is this assumption in practical scenarios? \\
\textsc{\footnotesize Insights} {\footnotesize\textit{nuance}} \newline \small How do the scaling parameters for |V| and p in your graph generation process affect the realism of the generated graphs compared to real-world graphs, and was any dataset used as a benchmark for realism? & \textsc{\footnotesize Insights} {\footnotesize\textit{more}} \newline \small Could you provide more details on the types of observational data that PINNs can integrate and how this impacts their generalization to different PDE scenarios? \\
\textsc{\footnotesize Method} {\footnotesize\textit{detail}} \newline \small How does the model determine which binding sites to predict as 'multiple binding sites,' and are there any criteria or thresholds involved in this prediction process? & \textsc{\footnotesize Insights} {\footnotesize\textit{soundness}} \newline \small Why were 3 random seeds used for RPM but 4 random seeds used for baselines? How does this difference affect the fairness of comparisons? \\
\textsc{\footnotesize Insights} {\footnotesize\textit{nuance}} \newline \small What specific parameters or features of the dataset could reflect the socio-political situation of the country, and how were these parameters selected? & \textsc{\footnotesize Insights} {\footnotesize\textit{nuance}} \newline \small How does augmenting the state with N in your approach compare quantitatively or qualitatively to the approach in Perrin et al. (2022)? Did you observe any differences in performance or generalization across games? \\
\textsc{\footnotesize Insights} {\footnotesize\textit{soundness}} \newline \small How is the spymaster's reward system calibrated to ensure fairness between varying levels of AI difficulty or human solvability? & \textsc{\footnotesize Method} {\footnotesize\textit{detail}} \newline \small How were the two obstacles placed at random locations? Was there any restriction on their placement, such as avoiding overlap with preys, predators, or each other? \\
\textsc{\footnotesize Insights} {\footnotesize\textit{more}} \newline \small Can you provide more precise details on how GNS outperforms other scheduling mechanisms when used with SGD, particularly in terms of specific metrics or cases? & \textsc{\footnotesize Method} {\footnotesize\textit{alternative}} \newline \small Why is the average of annotations chosen as an approximation of the soft ground truth label lx, and how do alternative aggregation methods compare to this approach in capturing aleatoric uncertainty? \\
\textsc{\footnotesize Insights} {\footnotesize\textit{nuance}} \newline \small How do the terms involving ($\beta$\_i - $\beta$\_i)($\alpha$\_i - $\alpha$\_i)O(log\textasciicircum{}-1 2 n) affect the asymptotic behavior of these probability bounds, and are there conditions under which they could become significant? & \textsc{\footnotesize Method} {\footnotesize\textit{alternative}} \newline \small Why did you choose to focus on roto-translation invariant features as opposed to exploring potential roto-translation equivariant features that might retain more input variability? \\
\textsc{\footnotesize Insights} {\footnotesize\textit{more}} \newline \small Could you elaborate on the specific mechanism or rationale behind the diminishing boost of synthetic data as the real shot number increases? Is it due to a saturation effect or some other factor? & \textsc{\footnotesize Insights} {\footnotesize\textit{more}} \newline \small What specific types of new losses are envisioned or recommended for targeting models robust to distributional shift beyond Generalized Reweighting approaches? \\
\textsc{\footnotesize Measurement} {\footnotesize\textit{alternative}} \newline \small Could an alternative choice for the initialization batch size (Binit) significantly alter the communication complexity or number of stochastic gradient calculations? If so, how? & \textsc{\footnotesize Method} {\footnotesize\textit{alternative}} \newline \small Why was the specific task-heterogeneous batching strategy used in this study preferred over other batching approaches, such as task-homogeneous batching or dynamic batching? \\
\textsc{\footnotesize Insights} {\footnotesize\textit{soundness}} \newline \small Could you provide a detailed justification for why the terms |Tj - Tj1| and |$\nu$ - $\nu$| are bounded by their respective expressions, and are there scenarios where these bounds become loose? & \textsc{\footnotesize Method} {\footnotesize\textit{alternative}} \newline \small Could alternative methods, such as transfer learning from Poker or Go datasets, have been considered instead of relying solely on expert demonstrations? If so, why was this not done? \\
\midrule
\end{longtable}

\section{Question Style Example}

We do notice that questions generated by 4o models repeat more of the content in the paragraph that it was generated for, which can be seen in Table~\ref{tab:copying_table}. We later quantify this observation in Section~\ref{sec:copying} and show that it holds in general, with higher percentage of words in the question drawn from the originating paragraph for 4o questions compared to o3 questions.

\newcolumntype{Q}{>{\raggedright\arraybackslash}X}
 
\newcommand{\commentary}[1]{
  \vspace{4pt}
  \begin{mdframed}[backgroundcolor=gray!12, linewidth=0pt, innerleftmargin=6pt, innerrightmargin=6pt, innertopmargin=4pt, innerbottommargin=4pt]
  \textit{#1}
  \end{mdframed}
}

\begin{table*}[ht]
\centering
\small
\renewcommand{\arraystretch}{1.3}
\begin{tabularx}{\textwidth}{@{} l Q @{}}
\toprule
 
\textbf{PeerQA}
  & \textbf{Q:} Do you evaluate playing strength of agents by restricting them by MCTS
    iteration counts or by time limits?
    
    \\
\midrule
 
\textbf{4o segment}
  & \textbf{Q:} What motivates the choice to approximate the number of MCTS simulations
    ($T$) as the maximum number allowed, and could this approximation
    influence results in scenarios with substantial late-game positions where
    the full game tree may already be mapped? \newline
    \textbf{Originating Span:} ``We also approximate the number of MCTS simulations $T$ to be
    the maximum number of simulations allowed, since the maximum is reached
    at all game positions except late-game positions, where the remaining
    game tree is already fully mapped.''
    
    \\
\midrule
 
\textbf{o3 full}
  & \textbf{Q:} You approximate $T$ by the maximum number of MCTS simulations allowed
    because that maximum is supposedly reached at almost all positions.
    Can you quantify what fraction of positions actually hit the maximum
    in practice and how deviations from the maximum affect the computed
    FLOPs and the fitted $\alpha_C$? \newline
    \textbf{Originating Span:} ``We also approximate the number of MCTS simulations $T$ to be
    the maximum number of simulations allowed, since the maximum is reached
    at all game positions except late-game positions, where the remaining
    game tree is already fully mapped.''
    
    \\
 
\bottomrule
\end{tabularx}
\caption{Table comparing similar questions asked by three different systems}
\label{tab:copying_table}
\end{table*}

\section{Example edit-inducing questions and respective edit rates}
\label{app:editRateExamples}

 Example edit-inducing questions and edit rates, along with the textual changes corresponding to these edit rates. All questions are human-written and sourced from the PeerQA dataset. 
\begin{table*}[t!]
\centering
\small
\renewcommand{\arraystretch}{1.3} 
\begin{tabularx}{\textwidth}{@{} >{\raggedright\arraybackslash\hsize=0.25\hsize}X >{\raggedright\arraybackslash\hsize=0.75\hsize}X @{}}
\toprule

\textbf{Type:} High Edit Rate \newline
\textbf{Edit Rate:} 0.617 \newline

&
\textbf{Q:} What are the global convergence criteria used when solving PESNet for a continuous subset of structures? \newline
\textbf{Initial:} For H$^+_4$ and cyclobutadiene, we train on discrete sets of geometries from the literature (Scherbela et al., 2021; Kinal \& Piecuch, 2007). \newline
\textbf{Final:} For H$^+_4$ and cyclobutadiene... \hl{[New Paragraph]... To still access convergence, we use the fact that the local energy $E_L$ of any eigenfunction (including the ground-state) has 0 variance...} \\
\midrule

\textbf{Type:} Low Edit Rate \newline
\textbf{Edit Rate:} 0.247 \newline

&
\textbf{Q:} What is the relationship between the $p^*()$ label and the one-hot encoding for the hard sample in Figure 3? \newline
\textbf{Initial:} we deﬁne base difﬁculty as $\|e_y - p^*(x)\|_2$. This will be high for ambiguous points, but especially points where the sampled $y$ had low probability under $p^*$. \newline
\textbf{Final:} we deﬁne base difﬁculty as $\|e_y - p^*(x)\|_2$, which is large if: $\bullet$ $x$ is ambiguous: \hl{$p^*$ has several large components, so there is no one-hot label near $p^*$.} \\
\midrule

\textbf{Type:} Invalid \newline

&
\textbf{Q:} How was the fine tuning done for the step sizes in the experiments? \newline
\textbf{Initial:} except for the step sizes -- we ﬁne-tune them using a set of powers of two $\{2^i \mid i \in [-10, 10]\}$ -- \newline
\textbf{Final:} except for the step sizes -- we ﬁne-tune them using a set of powers of two $\{2^i \mid i \in [-10, 10]\}$ -- \\
\midrule

\textbf{Type:} Unanswered \newline

&
\textbf{Q:} Is there a plan to open-source the proprietary medical knowledge base and the telemedicine software? \newline
\textbf{Initial:} \textit{nil} \newline
\textbf{Final:} \textit{nil} \\

\bottomrule
\end{tabularx}
\caption{Examples drawn from human written questions in the PeerQA dataset highlighting the different types of questions and how they are assessed.}
\label{tab:qualitative_examples}
\end{table*}

\section{License Information}

We used a subset of the PeerQA dataset~\cite{baumgärtner2025peerqascientificquestionanswering}, consisting of papers and their associated peer review questions sourced from OpenReview. Both the papers and peer reviews are publicly accessible via OpenReview under OpenReview's terms of use; individual copyrights remain with their respective authors and venues. Our use is solely for non-commercial research purposes.

\section{Hyperparameter Details}

We conduct all experiments using OpenAI's API. Specifically, we use \texttt{gpt-4o (gpt-4o-2024-11-20)} and \texttt{o3 (o3-2025-04-16)}. Both models are proprietary and their exact parameter counts are not publicly disclosed by OpenAI. All experiments use default hyperparameters (temperature = 1.0, top-p = 1.0) as provided by the API, with no custom decoding configuration.

\end{document}